\documentclass[final]{cvpr}
\usepackage{mathrsfs}

\usepackage{times}
\usepackage{epsfig}
\usepackage{graphicx}
\usepackage{amsmath}
\usepackage{amssymb}
\usepackage{comment}
\usepackage{bm}
\usepackage{multirow}

\usepackage[pagebackref=true,breaklinks=true,colorlinks,bookmarks=false]{hyperref}
\def\etal{\emph{et al}\onedot}

\def\cvprPaperID{7128} 
\def\confYear{CVPR 2021}

\newcommand{\snote}[1]{\textcolor{red}{\textbf{SK: #1}}}

\newcommand{\ynote}[1]{\textcolor{blue}{\textbf{YG: #1}}}

\begin{document}

\title{A Peek Into the Reasoning of Neural Networks: Interpreting with Structural Visual Concepts}

\author{Yunhao Ge$^{1,2}$, Yao Xiao$^{2}$, Zhi Xu$^{2}$, Meng Zheng$^{1}$, Srikrishna Karanam$^{1}$,\\ Terrence Chen$^{1}$, Laurent Itti$^{2}$, and Ziyan Wu$^{1}$\\
$^{1}$United Imaging Intelligence, Cambridge MA\\
$^{2}$ University of Southern California, Los Angeles CA \\
{\tt \scalebox{.7}{\{first.last\}@united-imaging.com,yunhaoge@usc.edu,yxiao915@usc.edu,zhix@usc.edu,itti@usc.edu}}
}

\maketitle

\pagestyle{empty}

\begin{abstract}
Despite substantial progress in applying neural networks (NN) to a wide variety of areas, they still largely suffer from a lack of transparency and interpretability. While recent developments in explainable artificial intelligence attempt to bridge this gap (e.g., by visualizing the correlation between input pixels and final outputs), these approaches are limited to explaining low-level relationships, and crucially, do not provide insights on error correction. In this work, we propose a framework (VRX) to interpret classification NNs with intuitive structural visual concepts. Given a trained classification model, the proposed VRX extracts relevant class-specific visual concepts and organizes them using structural concept graphs (SCG) based on pairwise concept relationships. By means of knowledge distillation, we show VRX can take a step towards mimicking the reasoning process of NNs and provide logical, concept-level explanations for final model decisions. With extensive experiments, we empirically show VRX can meaningfully answer ``why" and ``why not" questions about the prediction, providing easy-to-understand insights about the reasoning process. We also show that these insights can potentially provide guidance on improving NN's performance.

\end{abstract}
\section{Introduction}
\thispagestyle{empty}
With the use of machine learning increasing dramatically in recent years in areas ranging from security \cite{buczak2015survey} to medicine \cite{shen2017deep}, it is critical that these neural network (NN) models are transparent and explainable as this relates directly to an end-user's trust in the algorithm \cite{goodman2017european,adadi2018peeking}. Consequently, explainable AI (xAI) has emerged as an important research topic with substantial progress in the past few years.
\begin{figure}[t]
\begin{center}
\includegraphics[width=\linewidth]{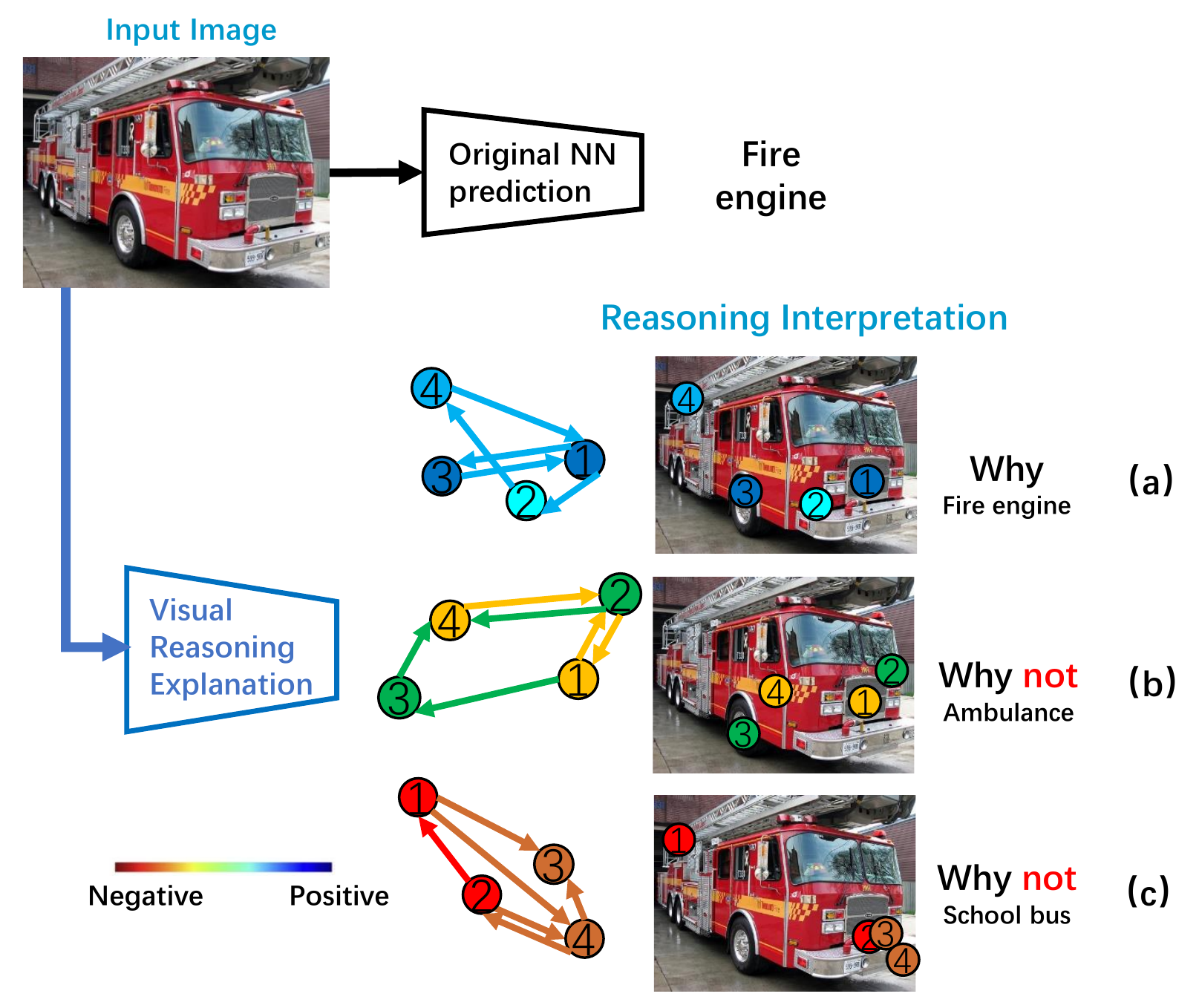}
\end{center}
   \caption{An example result with the proposed VRX. To explain the prediction (i.e., fire engine and not alternatives like ambulance), VRX provides both visual and structural clues. Colors of visual concepts (numbered circles) and structural relationships (arrows) represent the positive or negative contribution computed by VRX to the final decision (see color scale inset). (a): The four detected concepts (1-engine grill, 2-bumper, 3-wheel, 4-ladder) and their relationships provide a positive contribution (blue) for fire engine prediction. (b, c): Unlike (a), the top 4 concepts, and their relationships, for ambulance/school bus are not well matched and contribute negatively to the decision (green/yellow/red colors).}
\label{fig:why}
\end{figure}
Most recent xAI approaches attempt to  explain NN decision reasoning process with visualizations depicting the correlation between input pixels ( or low-level features) and the final output \cite{zeiler2014visualizing,mahendran2015understanding,CAM_CVPR16,residualAtt_CVPR17,SelGradCAM_ICCV17,AttentionIA_NIPS17, TCAV_ICML2018, gradCAMpp_WACV18, sundararajan2017axiomatic,ribeiro2016model}, with perturbation-based \cite{sundararajan2017axiomatic,ribeiro2016model} and gradient-based \cite{SelGradCAM_ICCV17, gradCAMpp_WACV18} methods receiving particular attention in the community. Despite impressive progress, we identify some key limitations of these methods that motivate our work. First, the resulting explanations are limited to low-level relationships and are insufficient to provide in-depth reasoning for model inference. Second, these methods do not have systematic processes to verify the reliability of the proposed model explanations \cite{kim2018interpretability,ghorbani2019interpretation}. Finally, they do not offer guidance on how to correct mistakes made by the original model.

We contend that explaining the underlying decision reasoning process of the NN is critical to addressing the aforementioned issues. In addition to providing in-depth understanding and precise causality of a model's inference process, such a capability can help diagnose errors in the original model and improve performance, thereby helping take a step towards building next-generation human-in-the-loop AI systems. To take a step towards these goals, we propose the visual reasoning explanation framework (VRX) with the following key contributions:

\begin{itemize}
    \item To understand what an NN pays attention to, given an input image, we use high-level category-specific visual concepts and their pairwise relationships to build structural concepts graphs (SCGs) that help to highlight spatial relationships between visual concepts. Furthermore, our proposed method can in-principle encode higher-order relationships between visual concepts.
    \item To explain an NN's reasoning process, we propose a GNN-based graph reasoning network (GRN) framework that comprises a distillation-based knowledge transfer algorithm between the original NN and the GRN. With SCGs as input, the GRN helps optimize the underlying structural relationships between concepts that are important for the original NN's final decision, providing a procedure to explain the original NN. 
    \item Our proposed GRN is designed to answer interpretability questions such as \texttt{why} and \texttt{why not} as they relate to the original NN's inference decisions, helping provide systematic verification techniques to demonstrate the causality between our explanations and the model decision. We provide qualitative and quantitative results to show efficacy and reliability.
    \item As a useful by-product, in addition to visual reasoning explanations, our method can help take a step towards diagnosing reasons for any incorrect predictions and guide the model towards improved performance.
\end{itemize}

\begin{figure*}[h!]
\begin{center}
\includegraphics[width=\linewidth]{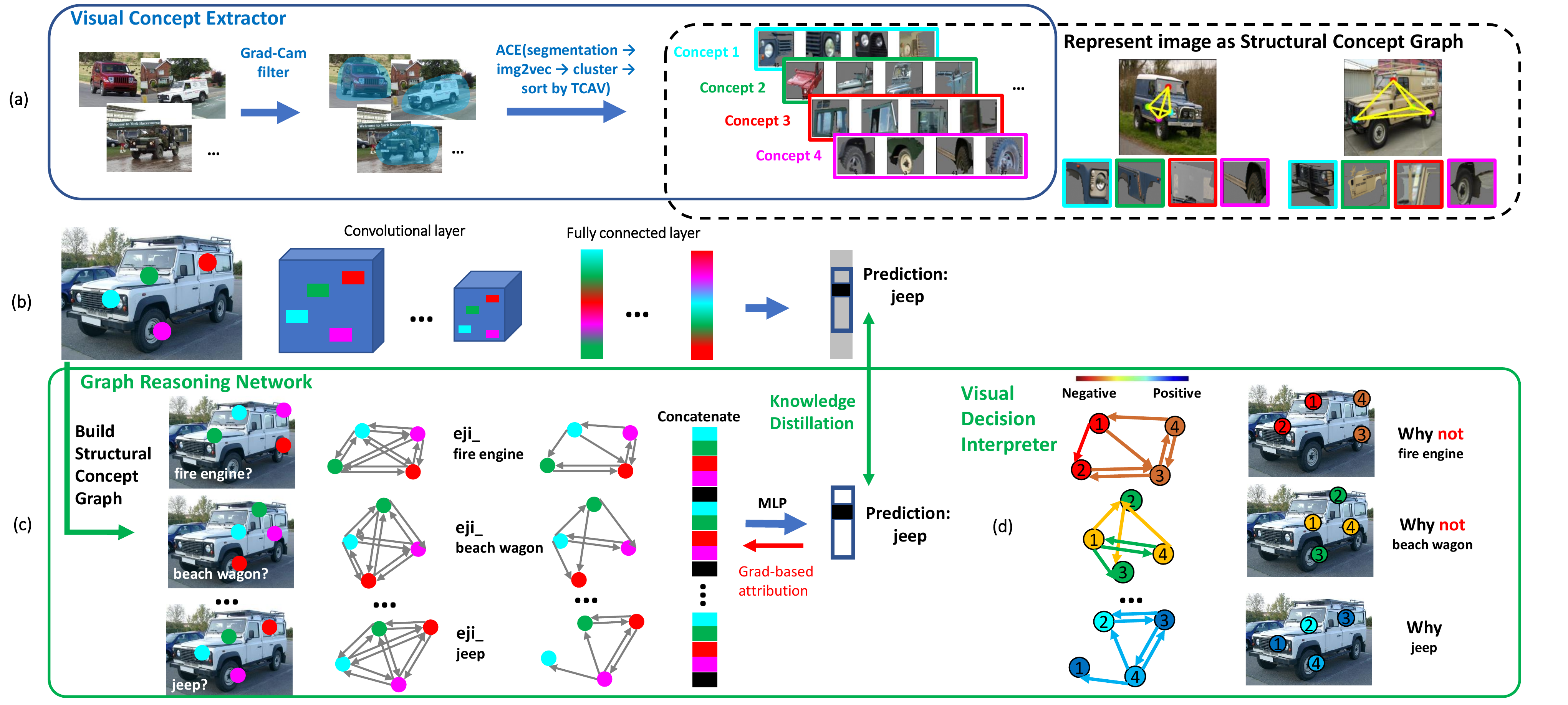}
\end{center}
   \caption{Pipeline for Visual Reasoning Explanation framework. (a) The Visual Concept Extractor (VCE) discovers the class-specific important visual concepts. (b) In original NN, the representation of the top $N$ concepts is distributed throughout the network (colored discs and rectangles). (c) Using Visual Concept Graphs that are specific to each image class, our VRX learns the respective contributions from visual concepts and from their spatial relationships, through distillation, to explain the network's decision. (d) In this example, the concept graphs colored according to contributions from concepts and relations towards each class explain why the network decides that this input is a Jeep and not others. }
\label{fig:2}
\end{figure*}







\section{Related Work}
In this section, we review existing literature relevant to our work interpreting convolutional neural networks, graph neural networks, and knowledge distillation to differentiate our method from others.

\noindent{\bf Interpreting neural networks.}
The substantial recent increase in the practical adoption of deep learning has necessitated the development of explainability and interpretability methods for neural networks (NNs), and convolutional neural networks (CNNs) in particular. One line of work focuses on pixel-level interpretation \cite{SelGradCAM_ICCV17, gradCAMpp_WACV18, CAM_CVPR16,ABN_CVPR19,GAIN_CVPR18,zheng2019re,wang2019sharpen}, producing attention maps to highlight the relevant image regions contributing to the final model decision. These methods can further be categorized into gradient-based and response-based methods. Response-based approaches use an additional computational unit to calculate the importance score of spatial image locations. For example, CAM \cite{CAM_CVPR16} utilized an auxiliary fully-connected layer to produce the spatial attention map and highlight image pixels contributing to the network decision. On the other hand, gradient-based methods, e.g., Grad-CAM \cite{SelGradCAM_ICCV17}, generate class-specific attention maps based on gradients backpropagated to the last convolutional layer given the model prediction. In addition to pixel-level interpretation, several recent works proposed to extract more human-intuitive concept-level explanations for interpreting neural networks \cite{TCAV_ICML2018, ACE_NIPS2019}. Specifically, Kim \etal \cite{TCAV_ICML2018} proposed TCAV where directional derivatives are used to quantify the sensitivity of the network's prediction with respect to input user-defined concepts. Ghorbani \etal proposed an automatic concept selection algorithm \cite{ACE_NIPS2019} based on the TCAV scores to produce meaningful concept-level explanations. 
While our framework also produces concept explanations automatically, it goes beyond this and learns explicit inter-concept relationships, producing more insightful interpretations.

\noindent{\bf Graph Networks.}
Graph neural networks (GNNs) have been successfully applied to tasks ranging from node classification \cite{kipf2016semi,hamilton2017inductive,xu2018representation}, edge classification \cite{perozzi2014deepwalk,grover2016node2vec} to graph classification \cite{gilmer2017neural, chen2018rise}. Based on ``message passing", powerful extensions such as GCNs \cite{kipf2016semi}, graph attention network (GAT) \cite{velivckovic2017graph}, SAGE \cite{hamilton2017inductive} and $k$-GNNs \cite{morris2019weisfeiler} have been proposed. Due to their trackable information-communication properties, GNNs can also be used for reasoning tasks, such as VQA \cite{teney2017graph,norcliffe2018learning} and scene understanding \cite{li2017scene}. In this work, we adopt the GCN to learn semantic relationships and interactions between human-interpretable concepts, providing more thorough explanations.

\noindent{\bf Knowledge distillation.}
Knowledge distillation can effectively learn a small student model from a large ensembled teacher model \cite{hinton2015distilling}, which finds broad applications in different areas, like model compression \cite{polino2018model} and knowledge transfer \cite{phuong2019towards}. In a similar spirit, in this work, we learn an easy-to-understand graph reasoning network (GRN) that produces the same classification decisions as the original NN model while also learning structural relationships between concepts to generate in-depth explanations for the original NN inference decisions.

\section{Visual Reasoning Explanation Framework}
\label{sec:3}
Our proposed visual reasoning explanation framework (VRX) to explain the underlying decision reasoning process of a given NN is visually summarized in Fig.~\ref{fig:2}. VRX comprises three main components: a visual concept extractor (VCE) to identify primitive category-specific visual concepts from the given neural network; a graph reasoning network (GRN) to organize category-specific visual concepts, represented as structural concept graphs (SCGs), based on their structural relationships, to mimic the decision of the original NN with knowledge transfer and distillation; and a 
visual decision interpreter (VDI) to visualize the reasoning process of the neural network given a certain prediction. We next explain each of these components in detail.

\subsection{Visual Concept Extractor}
\label{sec:3-1}

While most existing neural network explanation techniques focus on producing low-level saliency maps, these results may be suboptimal as they may not be intuitive for human users to understand. Inspired by the concept-based explanations (ACE) technique \cite{ACE_NIPS2019}, we propose to use visual concepts to represent an input image given class-specific knowledge of the trained neural network to help interpret its underlying decision-making processes. 

\begin{figure}[t]
\begin{center}
\includegraphics[width=\linewidth]{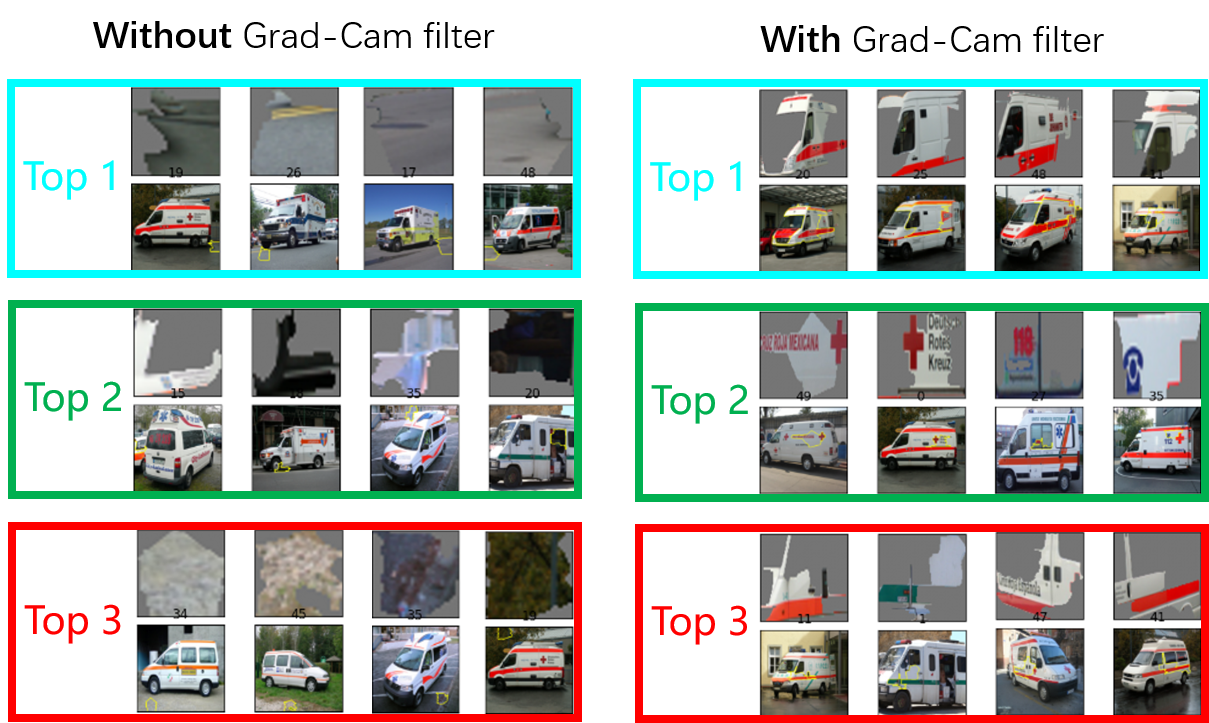}
\end{center}
   \caption{Concept discovery with and without Grad-Cam filter.}
\label{fig:filter}
\end{figure}

While ACE \cite{ACE_NIPS2019} is reasonably effective in extracting class-specific visual concepts, its performance is dependent on the availability of sufficient image samples for the given class of interest. As we show in Figure \ref{fig:filter} (left), for a class (ambulance here) with a small number of training images (50), the ACE concepts mostly fall on the background region, presenting challenges for a downstream visual explanation. To alleviate this issue, given an image $I$, we propose to use top-down gradient attention \cite{SelGradCAM_ICCV17} to first constrain the relevant regions for concept proposals to the foreground segments, thereby helping rule out irrelevant background patterns. Given the class-specific attention map $M$, we use a threshold $\tau$ to binarize $M$ as $\Bar{M}$ (pixel values lower than $\tau$ set to 0, others set to 1), which is used to generate the masked image $\Bar{I} = I \times \Bar{M}$ ($\times$ is element-wise multiplication) for further processing. Specifically, following ACE, we extract the top-N visual concepts and their mean feature vectors for each class of interest using the original trained NN. Fig.~\ref{fig:filter} demonstrates the importance of the proposed gradient attention pre-filtering discussed above using top-3 visual concepts for the ambulance class (concepts with the pre-filtering focus more clearly on the foreground). 

\subsection{Graph Reasoning Network}
\subsubsection{Representing Images as SCGs}
\label{sec:3-2-1}
Given the aforementioned class-specific visual concepts (see Section~\ref{sec:3-1}), we represent images using structural concept graphs (SCGs), which, as input to our proposed graph reasoning network (GRN), helps learn structural relationships between concepts and produce visual explanations for the original NN. Specifically, given an image, we use multi-resolution segmentation to obtain image patches (also called concept candidates), as inputs to the original NN to compute patch features, and then match these features to the mean concept feature vectors derived above (from Section~\ref{sec:3-1}). For each class of interest, we construct an SCG with concepts/patches detected from the input image, based on the Euclidean distance between patch feature and mean concept feature. Specifically, if the Euclidean distance between image patch feature and mean concept feature is larger than a threshold $t$, we identify this patch as a detected concept. For undetected concepts, we use dummy node feature representation (all feature values equal to a small constant $\epsilon$), to ensure network dimension consistency. Note that we have $n$ SCGs generated for the same input image considering all $n$ classes of interest.


SCG is a fully connected graph $(V,E)$ with bidirectional edges where each node $v_i \in V$ represents one relevant visual concept. Each directed edge 
$\text{edge}_{ji} = (v_j, v_i) \in E$ has two attributes: 1) a representation of spatial structure relationship between nodes $\text{edge}_{ji}$, initialized with the normalized image locations $[x_{j},y_{j},x_{i},y_{i}]$ of the two visual concepts it connects and updated in each layer of GRN; 2) a measure of dependency $e_{ji}$ (a trainable scalar) between concepts $v_i$, $v_j$ (see Fig.~\ref{fig:2} (c) and Fig.\ref{fig:eij} for an overview). Such a design helps our framework not only discover human-interpretable visual concepts contributing to network prediction but also how their underlying interactions (with $e_{ji}$ capturing the dependencies) affect the final decision.



\subsubsection{Imitate the Reasoning Process of NN}
In addition to learning concept representations and capturing the structural relationship between visual concepts 
we also need to ensure the proposed GRN follows the same reasoning process as the original NN. Since we represent images as SCGs, this problem comes down to optimizing the GRN, with SCG inputs, so it gives the same output/prediction as the original NN with image inputs. We realize this with a distillation-based training strategy.

Specifically, given an input image $I$ and a trained NN classifier $\mathcal{F}(\cdot)$, along with $n$ SCG hypotheses $\mathbf{h} = \{h_1,h_2,...h_n\}$ extracted from the input image, we seek to learn the GRN $\mathcal{G}$ for $\mathbf{h}$ such that $\mathcal{G}(\mathbf{h}) = \mathcal{F}(I)$, i.e., ensuring prediction consistency between the GRN and the original NN.
The proposed $\mathcal{G}(\cdot)$ comprises two modules: 1) a GNN $G$ is applied for all classes with different class-specific $e_{ji}$ to learn the graph representation of SCGs; 2) an embedding network $E$ is used to fuse multi-category SCGs for final class prediction, i.e.:
\begin{equation}
   \mathcal{G}(\mathbf{h}) = E(G(\mathbf{h})) = \mathcal{F}(I)
\end{equation}
Fig. 2(b-c) give an overview of the component relationship between the original NN (b) and the proposed GRN (c), showing how GRN learns the ``embedding" for each hypothesis and through knowledge distillation ensures the same prediction as the original NN.

We use GraphConv \cite{morris2019weisfeiler} as $G$'s backbone network and modify the aggregate weights. For each graph convolutional layer, we have:
\begin{equation}
	f^{i}_{k+1}=W_{1}f^{i}_{k} + \sum_{j\in \mathcal{N}(i)}e^c_{ji}W_{2}f^{j}_{k}
	\label{eq:eij_1}
\end{equation}
where $f^{i}_{k}$ denotes the feature of node $v_{i}$ (representing a concept) in layer $k$, $W_{1}$ and $W_{2}$ denote the shared linear transformation parameters for center node $v_{i}$ and neighbor node $v_{j}$ respectively, $\mathcal{N}(i)$ denotes the neighboring node sets connected to node $i$, and $e^c_{ji}$ denotes the aggregation weight from start node $v_{j}$ to end node $v_{i}$ for a certain class $c$, indicating the inter-dependency of concepts $i$ on $j$. Instead of using shared edges for all classes of interest, GRN learns class-specific $e^{c}_{ji}$, i.e. different aggregation weights for different classes to capture varying structural relationships between class-specific concepts. 

In order to better capture inter-concept relationships, we concatenate edge features with neighboring node features, denoted as $\mathcal{C}(e^{c}_{ji}W_{2}f^{j}_{k}, \text{edge}_k^{ji})$, and Equation \ref{eq:eij_1} becomes:
\begin{equation}\label{Eq.3}
	f^{i}_{k+1}=W_{1}f^{i}_{k} + \sum_{j\in \mathcal{N}(i)} W_{3}\mathcal{C}(e^{c}_{ji}W_{2}f^{j}_{k}, \text{edge}_k^{ji})
\end{equation}
With $\text{edge}^{ji}_{k+1}=W_{4}\text{edge}^{ji}_{k}$, and $W_{3}$ and $W_{4}$ denoting one layer linear transformation for concatenated message feature and edge feature respectively. Since $e^{c}_{ji}$ is a trainable parameter by design in our $G$, it helps learn concept inter-dependency as measured by the overall training objective (see Fig.~\ref{fig:eij}(b) for a fire engine image example). 

\begin{figure}[t]
\begin{center}
\includegraphics[width=0.7\linewidth]{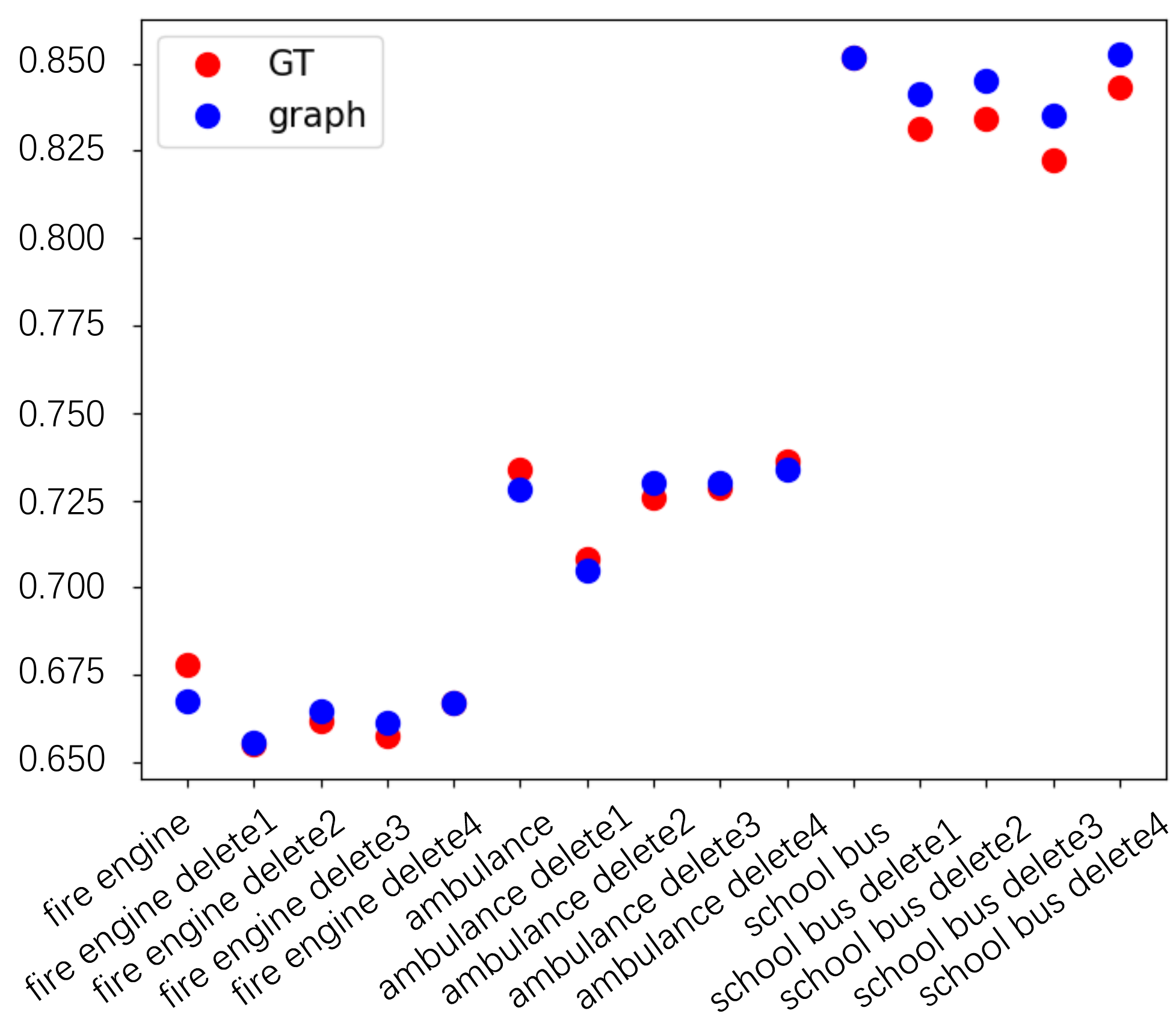}
\end{center}
   \caption{Decision comparison between original NN and proposed GRN.}
\label{fig:trainig}
\end{figure}
\begin{figure}[t]
\begin{center}
\includegraphics[width=\linewidth]{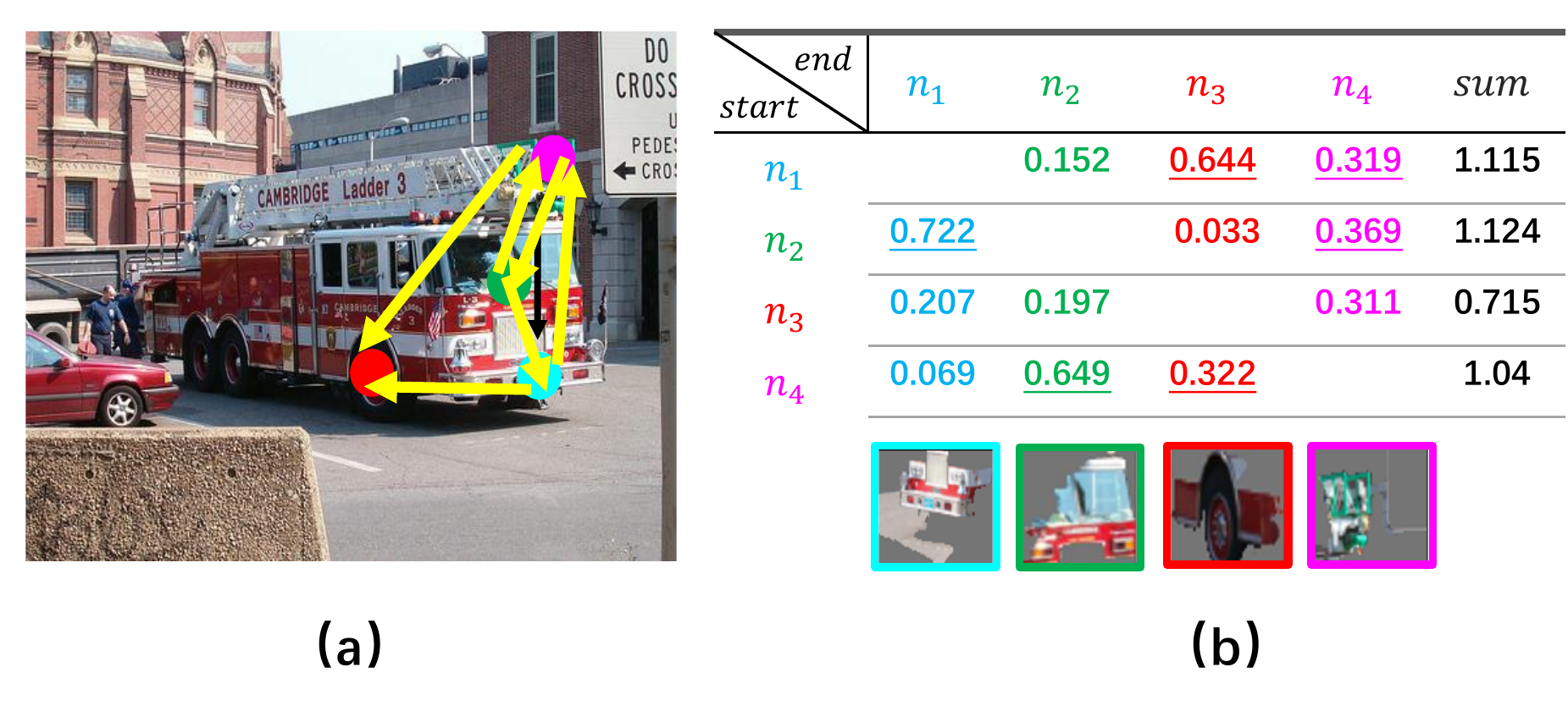}
\end{center}
   \caption{(a) Class-specific importance weights $e_{ji}$ highlight the important concept relationships for different classes (b) $e_{ji}$ reveals the information transformation between concepts, which shows the dependency between concepts: concept 1 and 2 contribute most information to other concepts, which makes them the 2 most discriminating concepts for a fire engine. }
\label{fig:eij}
\end{figure}

The embedding network $E$ concatenates all the feature vectors output from $G$ and maps it into a $n-$dimensional vector with an MLP ($n$ is the number of classes of interest). The GRN is then trained to imitate the original NN (see Fig.~\ref{fig:trainig}) by minimizing:
\begin{equation}
    \mathcal{L}_\textrm{d} =  
    || \sigma(\mathcal{G}(\mathbf{h})) - \sigma(\mathcal{F}(I)) ||_{l_1}
\end{equation}
where $\sigma(\cdot)$ is a normalization function (see Supplementary for more implementation details).
To imitation robustly, we randomly mask out one of the detected visual concepts on the input image. Fig.~\ref{fig:trainig} demonstrates the prediction comparison between the learned $\mathcal{G}$ and the original NN. \{\emph{class name}\}\textunderscore detect\{\emph{N}\} denotes images from category \emph{class name} with concept \emph{N} masked out.

\begin{figure*}[t!]
\begin{center}
\includegraphics[width=\linewidth]{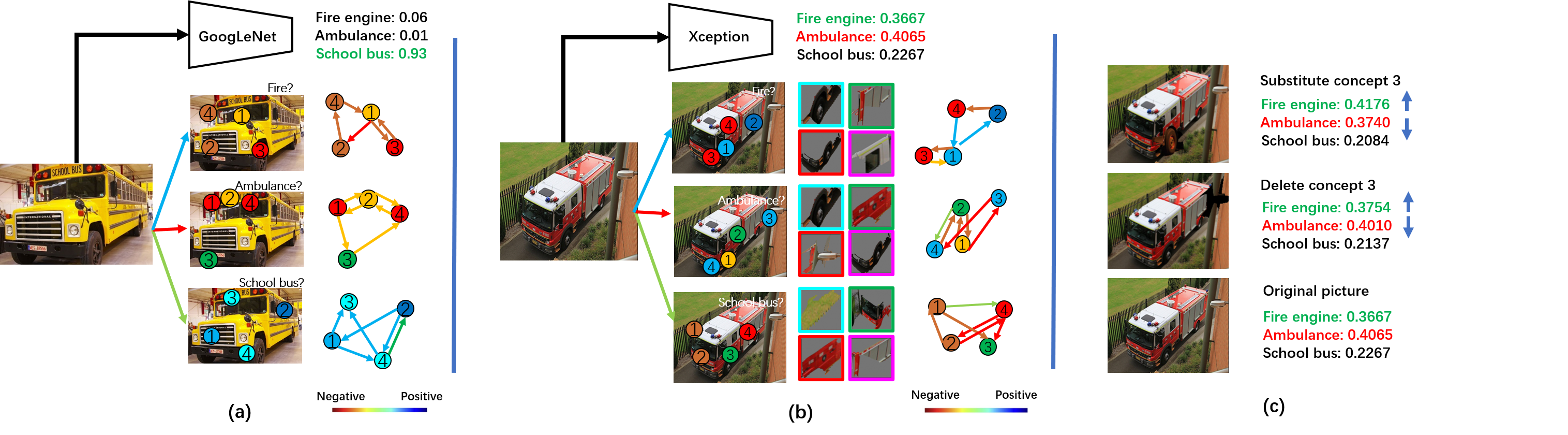}
\end{center}
   \caption{Visual Reasoning Explanation and logic consistency experiment example.}
\label{fig:ex_ex}
\vspace{-7pt}
\end{figure*}
\subsection{Visual Decision Interpreter}
\label{sec:3-3}

Once our GRN is trained to be a structural-concept-level representation of the original neural network, we can then interpret the original model decisions with our visual decision interpreter (VDI) module. As shown in Fig.~\ref{fig:2}(c-d), after feeding an image to both the original NN and the GRN, 
we obtain the final prediction $y$ representing the probability of all class of interest, $y = E(G(\mathbf{h}))=E(\mathcal{C}^m_{i=1}(G^{i}(h_i)))$. where $G^{i}$ represents the shared $G$ equipped with class $i$'s aggregate weight $e^{i}_{ji}$ and $G^{i}(h_i)$ is the graph embedding for the $i$th hypothesis SCG composed of the extracted concept node and edge feature representations; $\mathcal{C}$ denotes concatenation operation. For each interested class $c$, we have a class prediction score $y^{c}$
and compute gradients of $y^{c}$ with respect to the graph embeddings from $m$ hypothesis as:
\begin{equation}\label{Eq.6}
\bm{\alpha}_i = \frac{\partial y^{c}}{\partial G^{i}(h_i)}, i={1,...,m}
\end{equation}
where $\bm{\alpha}_{i}$ denotes the contribution weight vector of hypothesis $h_i$. The contribution score $s_{i}$ for each hypothesis $h_{i}$ w.r.t the prediction of $y^{c}$ is computed as the weighted sum of $\bm{\alpha}_{i}$ and $G(h_i)$:
\begin{equation}\label{Eq.7}
s_i = \bm{\alpha}_{i}^T G^{i}(h_i), i={1,...,m}
\end{equation}


We then use the contribution score $s_{i}$ computed from Eq.~\ref{Eq.7} to
indicate the positive or negative contribution (contribution score) of each node (concept) or edge (spatial and dependency conceptual relationship) to the decision made by the neural network (positive contribution score means positive contribution and vice versa).

\section{Experiments and results}

We conduct four different experiments to demonstrate the effectiveness of our proposed VRX in interpreting the underlying reasoning logic of neural network's decision, guiding network diagnosis and improving the performance of the original neural network. In our experiments, we use Xception \cite{Xception_CVPR17} and GoogLeNet models \cite{szegedy2015going} pre-trained on the ILSVRC2012 dataset (ImageNet) \cite{deng2009imagenet} as the target neural networks. 

\subsection{Visual Reasoning Explanation Experiment}
Fig.~\ref{fig:ex_ex} (a-b) shows two examples (one correct and one incorrect prediction) of how our VRX can help to explain the decision behind neural networks by performing experiments on GoogLeNet and Xception, respectively. 

Given a pre-trained GoogLeNet on ImageNet, we develop a VRX as introduced in Sec.~\ref{sec:3} to explain the reasoning logic. As shown in Fig.~\ref{fig:ex_ex} (a), for the input school bus image, both GoogLeNet and our VRX correctly predict the input as a school bus, with VRX outputs nearly identical prediction vector as original GoogLeNet which aligns with our expectation that our VRX ideally should imitate the behavior of original NN. We then use our proposed VRX to compute the contribution score for each concept node and edge to analyze how the detected human-interpretable concepts along with their structural relationships contributing to the network's decision. In this case, we ask `why school bus?' (why the original NN predict this image as a school bus?): from a visual/conceptual perspective, all detected top 4 important concepts have high positive contribution (blue) to the prediction probability of school bus (Row 3 of Fig.~\ref{fig:ex_ex} (a)), indicating the network is able to discover meaningful visual regions contributive to the correct prediction; from a structural perspective, the spatial location and relationship between concepts represented by edge arrows also contribute positively (light or dark blue), meaning the network identifies correct spatial correlations between detected visual concepts. Similarly, to answer `why not fire engine?' and `why not ambulance?', VRX identifies nearly all detected concepts negatively contribute to the corresponding prediction class, and all structure relationships between concepts have negative contributions to the class prediction as well. Based on the explanation above, VRX can give a systematically in-depth and easy-to-understand interpretation of the decision-making logic of GoogLeNet, from the visual and structural perspectives respectively.

The second example is shown in Fig.~\ref{fig:ex_ex} (b) for Xception network. Given an image of a fire engine, both the original Xception and our VRX wrongly predict ambulance as output. To understand why original Xception makes the incorrect prediction, our VRX is able to provide both visual and structural clues as well. From Fig.~\ref{fig:ex_ex} (b) Row 1, we can see that the detected visual concepts 3 (wheels of the vehicle) and 4 have negative contribution to the prediction of fire engine class, indicating that the wheel region of the input image is not consistent with the model's knowledge of fire engine (with negative contribution). To answer "why ambulance", concept 3 and 4 have positive contribution to ambulance prediction, which explains why the original Xception network incorrectly predicts the input image as an ambulance.

\subsection{Logic Consistency between VRX and NN}

To verify that the explanation of VRX is logically consistent with the reasoning of Xception, we present two experiments as follows. First, as shown in Fig.~\ref{fig:ex_ex} (c), for the wrong prediction example same as Fig.~\ref{fig:ex_ex} (b), we substitute the flawed fire engine concept 3, which has a negative contribution (low contribution score), with a good concept 3 (high contribution score) from another fire engine image and form a new modified image. Then, we use Xception to re-predict the class of the modified image, it corrects the error and predicts the input as a fire engine correctly. To show a causal relationship between VRX's explanation and the reasoning logic of Xception, we perform two additional contrastive experiments: a) Random substitute: if we substitute concept 3 with random patches, Xception does not achieve a correct prediction; b) Substitute good: if we substitute concepts 1 or 2 with other equivalently good patches from other images of fire engines, Xception also does not produce a correct decision. Thus, we conclude that VRX has correctly diagnosed the cause of Xception's error (here, a bad concept 3). Below, we show how this can be used to further guide improved training of original NN without manually modifying the image. 


For the wrongly predicted class, ambulance, if we delete a concept patch with a high contribution to ambulance probability, the prediction of Xception shows a decreased probability prediction of ambulance class and a higher probability prediction of fire engine.
In total, we applied this experiment to 119 images that were initially wrongly predicted by Xception (Table.~\ref{tab-1}). The results show that with the guidance of VRX (confusing/bad concept detected), most wrong prediction cases can be corrected through learning-based modifications of the images. 

\begin{table}[t]
\footnotesize

\begin{center}
  \begin{tabular}{c|cccc}
\hline
      &  & \multicolumn{3}{c}{Cause of error}  \\
\hline
   Error type    &    total &   concept &  structure & both\\
 \hline

 Before correction  &  119 &  5 &  6 & 108\\
 \hline
   \hline
 Substitute with\\ Random patches &  117 &  5 &  6 & 106 \\
 \hline
 Change good concepts &  115 &  5 &  6 & 104 \\
  \hline
 \textbf{VRX guided correction} &  \textbf{5} &  \textbf{1} &  \textbf{2} &  \textbf{2}\\
  \hline
\end{tabular}

\end{center}
\caption{VRX model helps correction. Out of 119 images initially misclassified by Xception, only 5 remain misclassified after VRX-guided image editing. Over 30\% of the samples have missing concepts and over 95\% of them have been correctly explained. In contrast, 117 and 115 images remain misclassified after substituting bad concepts with random image patches, or substituting good concepts with other good concepts from other images from the same class.}
\label{tab-1}
\end{table}


\subsection{Interpretation Sensitive of Visual and Structure }
\begin{figure}[t]
\begin{center}
\includegraphics[width=\linewidth]{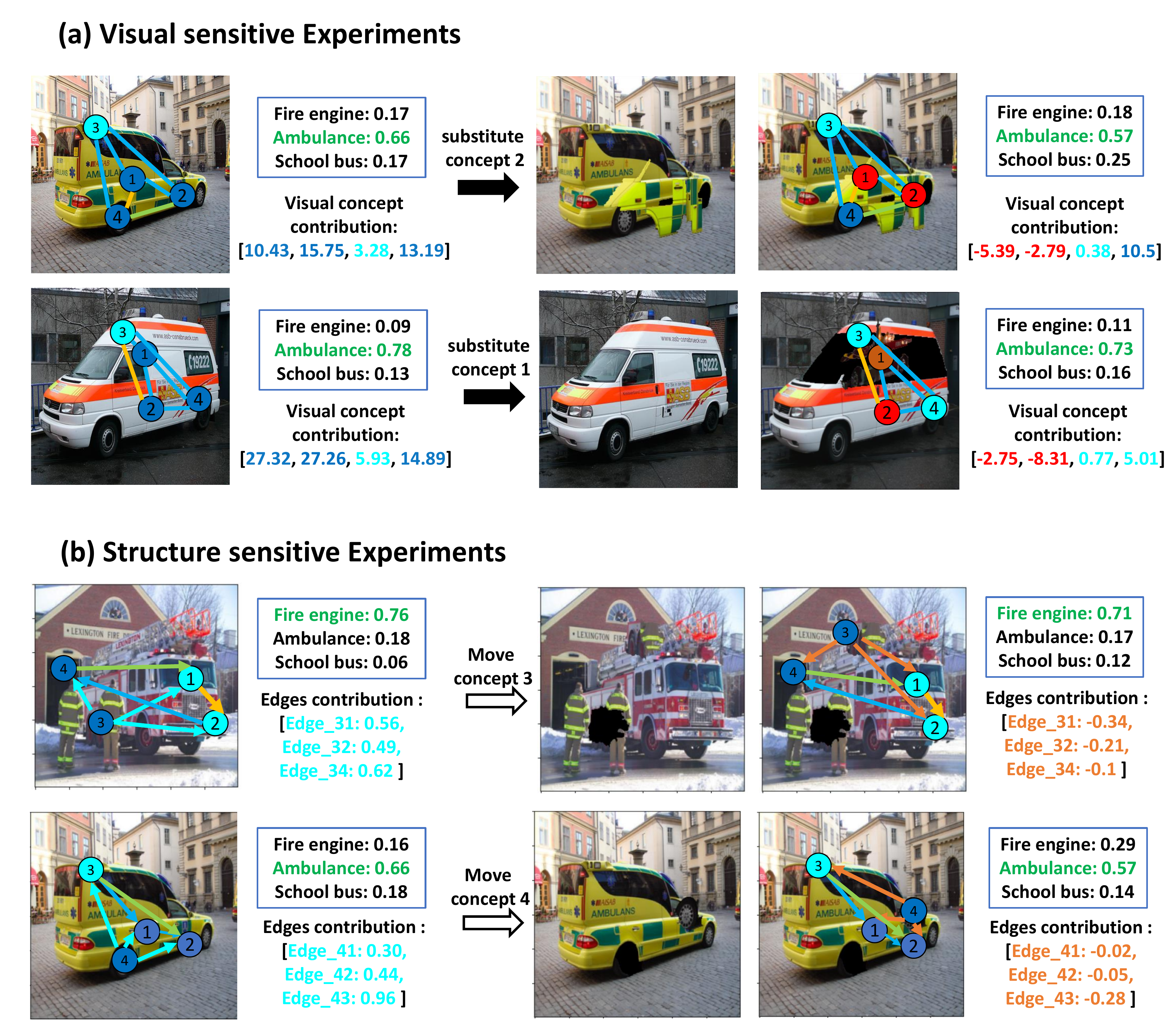}
\end{center}
   \caption{Interpretation from VRX is sensitive to visual and structure aspects. (a) visual sensitive (b) structure sensitive.}
\label{fig:sensitive}
\vspace{-7pt}
\end{figure}
We have demonstrated that VRX can help explain why and why not the model makes the decision, and shows a causal relationship between VRX's explanation and the original NN's decision. In this section, we focus on the sensitivity analysis of VRX's explanation from visual and structural aspects, respectively. We design two experiments accordingly: first, when we substitute a relatively good concept (with high positive contribution scores to corresponding class prediction) patch with a relatively bad concept (with lower positive or even negative contribution score) patch in an image, we want to see if VRX can capture the difference and precisely locate the correct modification, which shows the sensitivity of VRX to visual explanation. Second, when we move one concept's location from a reasonable place to an abnormal location, we want to make sure if VRX can precisely capture the structural abnormality and produce a corresponding explanation that correctly matches our modification.

Fig.~\ref{fig:sensitive}(a) demonstrates two visual sensitivity experiment examples. In the top row, given an ambulance image with a correct prediction from a trained Xception (Fig.~\ref{fig:sensitive}(a) left), VRX explains that all detected concepts and relative structure relationship have positive contributions to the prediction of ambulance class. We then substitute the original good concept 2 with relatively bad concept 2 from another ambulance image and form a modified ambulance image (Fig.~\ref{fig:sensitive}(a) right), to check the sensitivity of our VRX with respect to visual perturbation. From Fig.~\ref{fig:sensitive}(a), we can see that after the substitution, the class prediction score from both VRX and original Xception decrease as expected. While VRX gives a clear explanation for this performance decrease due to: less contributive concept 1 and 2 (negative contribution to the ambulance prediction), and invariant structure contributions, which correctly matches our modification in the original image. This proves the sensitivity of our VRX to visual perturbations. The second row of Fig.~\ref{fig:sensitive}(a) shows an additional example of visual sensitivity test.

Fig.~\ref{fig:sensitive}(b) illustrates two structure sensitivity experiments. Given a fire engine image with a correct prediction from trained Xception, VRX shows that concept 3 and the structural relationships of concept 3 to all adjacent concepts are positively contributive for class prediction. We then move concept 3 from the original location to an abnormal location (we move the wheels from the bottom to the sky) and form a modified fire engine image (Fig.~\ref{fig:sensitive}(b) right) to test the structural sensitivity of our VRX. Similarly, VRX produces consistent explanation with respect to structure perturbation as well, where the spatial relationship importance score between concept 3 to all adjacent concepts decrease after the substitution, which demonstrates the good sensitivity of our VRX to structural information. A second example in Fig.~\ref{fig:sensitive}(b) shows similar results.

\subsection{Model Diagnosis with VRX}

\begin{figure}[t]
\begin{center}
\includegraphics[width=\linewidth]{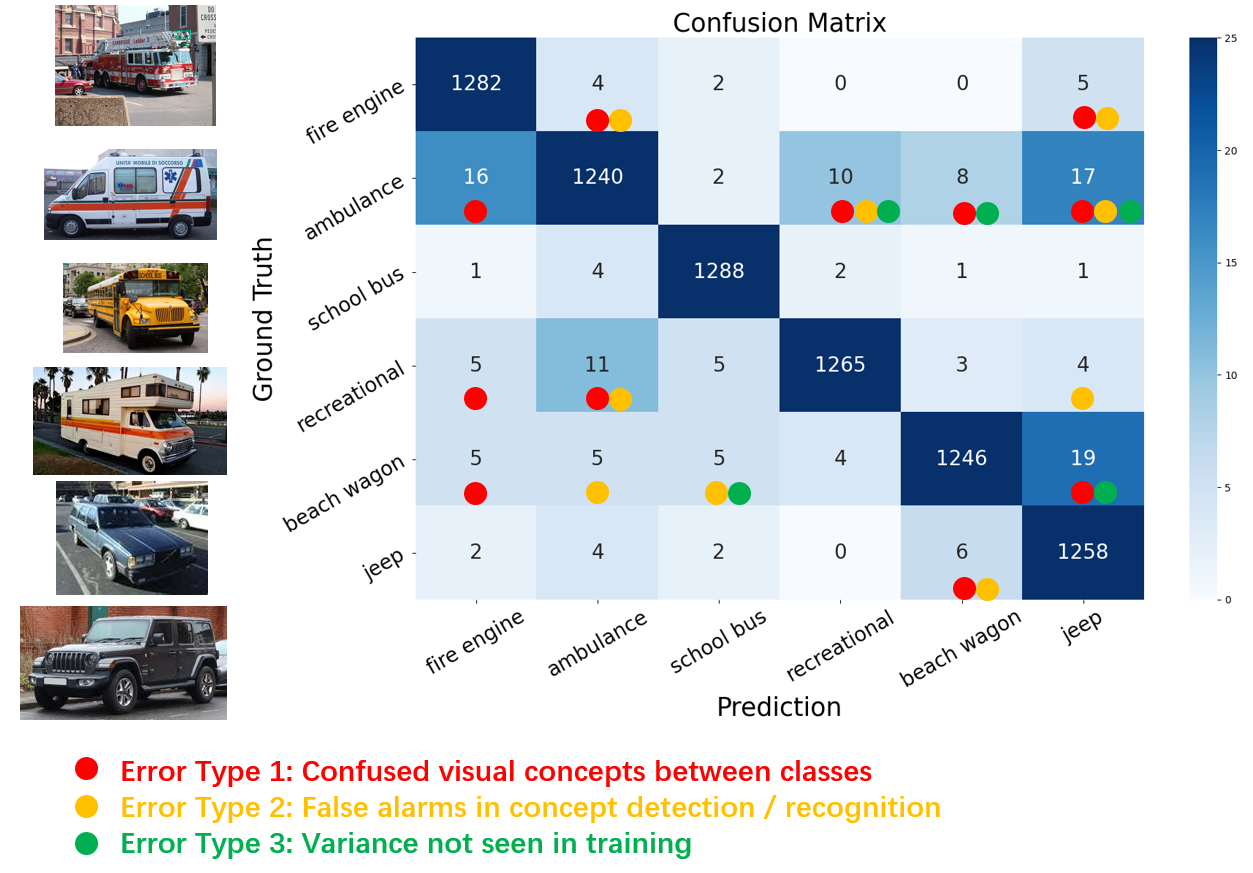}
\end{center}
   \caption{Model diagnosis and improving performance}
\label{fig:confu}
\vspace{-7pt}
\end{figure}
With the explainability of VRX, reasoning results generated by VRX can be further utilized to guide improving the performance and generalizability of the original NN. Fig.~\ref{fig:confu} shows a 6-class confusion matrix with Xception. With VRX, the type of error Xception makes can be categorized as the following:

(1) Confused visual concepts between classes. The top $k$ concepts of different classes may share certain overlaps. For instance, most vehicles have concepts related to 'wheels'. Hence judging only by this concept, the neural network may confuse one type of vehicle with another. There are existing approaches \cite{li2018tell} which can guide the network in growing its attentive region and alleviating the impact from biases in training data.  

(2) False alarms in concept detection/recognition. To VRX this usually means one or more patches are incorrectly labeled, which means either the neural network's feature extraction can be improved, or the most important visual concepts for specific classes are not discriminative enough. 

(3) Variance not seen in training. For instance, the distribution of viewpoints of a class of interest is biased in the training set of the NN. When the same object with an unseen viewpoint is presented to the NN, it may fail to recognize it. In these cases, in VRX's  decision reasoning, it may appear that most of the detected concepts are very close matches. However, the edge features seem off, suggesting the structural or spatial relationships between concepts are the cause for the NN to make incorrect predictions. 
Augmenting the training images with more diversity in viewpoints may solve the problem, as the further experiment shown below with the iLab-20M \cite{borji2016ilab} dataset.

\begin{figure}[t]
\begin{center}
\includegraphics[width=0.8\linewidth]{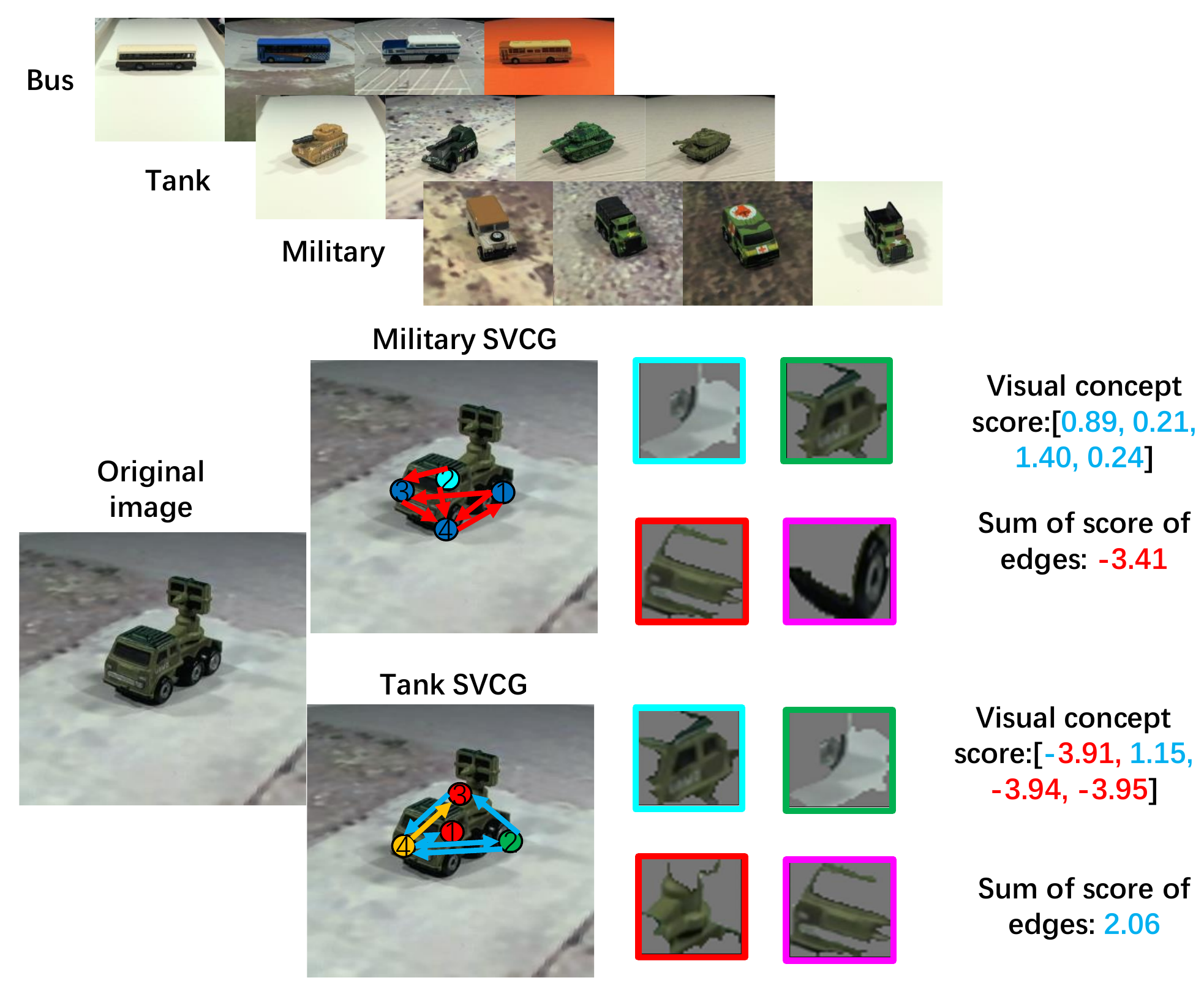}
\end{center}
   \caption{Diagnosis and improvement experiment on iLab-20M.}
\label{fig:ilab}
\vspace{-10pt}
\end{figure}

To further demonstrate the capability of NN diagnosis, we design an experiment on iLab-20M. iLab-20M is an attributed dataset with images of toy vehicles on a turntable captured with 11 cameras from different viewpoints.
We sampled a subset from iLab-20M with similar identity and pose: we focus on three classes of vehicles: bus, military, and tank. In the training set, each class has 1000 images. We manually introduce biases with the pose of each class: all buses are with pose 1, all military are with pose 2 and all tanks are with pose 3 (Fig.~\ref{fig:ilab}). We designed an unbiased test set where each kind of vehicle has all the 3 poses. 

We train a Resnet-18 \cite{he2016deep} to classify the 3 types of vehicles with the training set and test the accuracy on the test set (Table.~\ref{table:ilab}).
To explain the reasoning logic of the trained network, we trained a GRN with VRX and explained the logic of common mistakes made by the Resnet-18 (Details in supplementary). For most incorrectly classified samples in the test set, given the input image (in Fig.~\ref{fig:ilab}, the military is wrongly predicted as tank), VRX's interpretation shows that most of the detected visual concepts had a positive contribution to the correct class while the structure relationship between concepts contributed mostly negatively, which leads to the incorrect prediction. 
To verify the ``diagnosis", we designed a follow-up experiment, focusing on improving performance for the military class. Setting 1: we add images of additional poses (150 for each of the three poses) for the military in the training set and test the performance on the test set; setting 2: we add the same amount of images (450) as setting 1 but with images of the same pose as in the original training set. Table 2 shows that the accuracy with the augmented training set using setting 1 obtains much higher performance compared to the initial experiment and the follow-up experiment with setting 2 which does not bring any improvement. This suggests that VRX can help to diagnose the root cause of mistakes a neural network made, and potentially provide useful suggestions to improve the original NN's performance. 
\begin{table}[t]
\small
\begin{center}
\begin{tabular}{c|c|c|c}
\hline
        & original & setting 1 & setting 2  \\
\hline
\hline
Average accuracy & 50 & 60 & 50\\
\hline

\end{tabular}
\end{center}
\caption{Testing set accuracy comparison for VRX boost original model performance. All numbers are in \%. }
\label{table:ilab}
\vspace{-5pt}
\end{table}

\section{Conclusion}
We considered the challenging problem of interpreting the decision process of a neural network for better transparency and explainability. We proposed a visual reasoning explanation framework (VRX) which can extract category-specific primitive visual concepts from a given neural network, and imitate the neural network's decision-making process. Our experiments showed that the VRX can visualize the reasoning process behind neural network's predictions at the concept level, which is intuitive for human users. Furthermore, with the interpretation from VRX, we demonstrated that it can provide diagnostic analysis and insights on the neural network, potentially providing guidance on its performance improvement. We believe that this is a small but important step forward towards better transparency and interpretability for deep neural networks.

\vspace{-10pt}
\subsubsection*{Acknowledgments}
\vspace{-7pt}
This paper is based on the work done during Yunhao Ge's internship with United Imaging Intelligence. The research was partially supported by C-BRIC (one of six centers in JUMP, a
Semiconductor Research Corporation (SRC) program sponsored by DARPA),
the Army Research Office (W911NF2020053), and the Intel and CISCO
Corporations. The authors affirm that the views expressed herein are
solely their own, and do not represent the views of the United States
government or any agency thereof.

\clearpage
\appendix
\section*{Appendix}

\section{Implementation Details}

\textbf{Network structure} The network architectures of Graph Neural Network $G$ and Embedding Network $E$ are shown in Table.~\ref{tab:supp1}. $G$ takes $n$ hypotheses $\mathbf{h} = \{h_1,h_2,...h_n\}$ (each hypothesis $h_{i}$ is in the form of Structural Concept Graph (SCG)) as input, and output $n$ feature vectors ($G(h_{i})$) which concatenate all updated node and edge features of $h_{i}$. In $G$, we use class-specific $e^{c}_{ji}$ for different hypotheses in each GraphConv layer.  $E$ concatenates all $n$ feature vectors from all the hypotheses into a long vector and
maps the vector (1 $\times$ (188 $\times$ $n$)) into $n$ dimensional vector (1 $\times$ $n$) with a MLP, where $n$ is the number of classes of interest. ``node" denotes node feature, ``edge" denotes edge feature, ``GraphConv" is graph convolutional layer, ``ReLU" denotes ReLU activation function, ``BN" denotes batch normalization, and ``FC" denotes fully connected layer.

\begin{table}[h]
\footnotesize
\begin{center}
\begin{tabular}{p{0.1cm}p{3.6cm}p{3.4cm}}
\hline\noalign{\smallskip}
Part & Input $\rightarrow$ Output Shape & Layer Information\\
\noalign{\smallskip}
 \hline
 \hline
 \noalign{\smallskip}
 \multirow{6}*{$G$} & node:(2048$\rightarrow$64); edge:(4 $\rightarrow$ 5) & GraphConv-($e^{c}_{ji}$), ReLU, BN\\\noalign{\smallskip}
 ~ & \ \  node:(64 $\rightarrow$ 32); edge:(5 $\rightarrow$ 5) & GraphConv-($e^{c}_{ji}$), ReLU, BN\\\noalign{\smallskip}
 ~ & \ \  node:(32 $\rightarrow$ 32); edge:(5 $\rightarrow$ 5) & GraphConv-($e^{c}_{ji}$), ReLU, BN\\\noalign{\smallskip}

 \noalign{\smallskip}
 \hline
 \noalign{\smallskip}
 $E$  & (188 $\times$ $n$) $\rightarrow$ ($n$) & FC-(188 $\times$ $n$, \ \ $n$)\\
 \noalign{\smallskip}

\hline
\end{tabular}
\end{center}
\caption{Network architectures of Graph Neural Network $G$ and Embedding Network $E$.}
\label{tab:supp1}
\end{table}

\textbf{Training details} In Section 3.3 of the main paper, Eq.2 explains the knowledge distillation  we used to imitate the reasoning process of Xception on ImageNet dataset. We train $G$ and $E$ in a end-to-end manner. Below are the details: we use Adam with $\beta_{1}$=0.9 and $\beta_{2}$=0.999, batch size 128, learning rate 0.01 for the first 100 epochs and use a decay rate of 0.5 for the next 200 epochs. For each interested class, we use 400 images to train and other 900 images to test.

\section{Model Diagnosis with VRX Details}
\begin{figure}[t]
\begin{center}
\includegraphics[width=\linewidth]{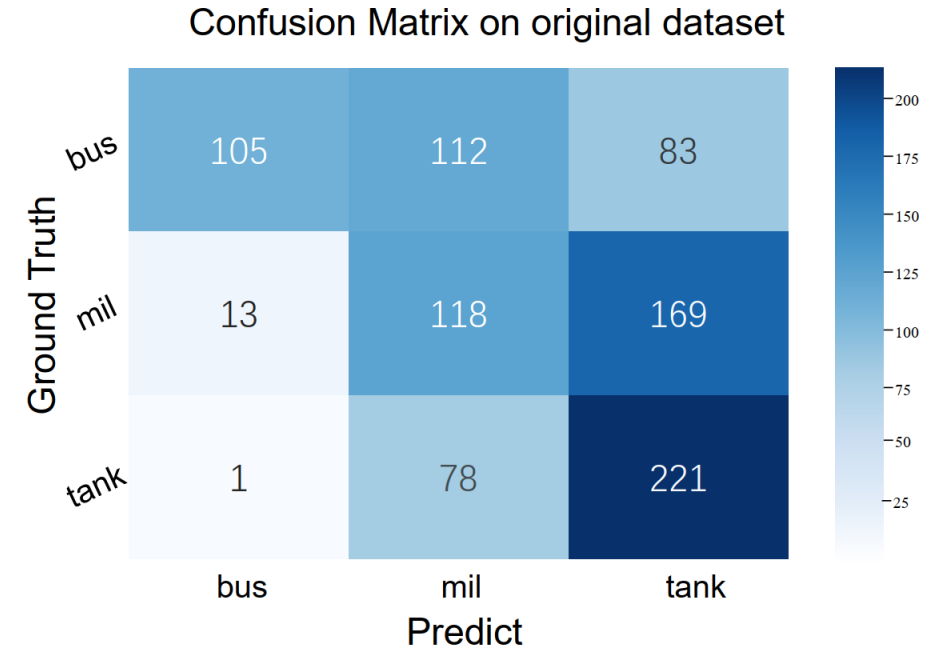}
\end{center}
   \caption{Confusion matrix of 3 class vehicle classification on test dataset with Resnet-18 trained on the original training set.}
\label{fig:sup-con}
\end{figure}

As we mentioned in mian paper Section 4.4,
Fig.~\ref{fig:sup-con} demonstrates the confusion matrix of 3 class vehicle classification on original dataset with Resnet-18. 

\end{document}